\documentclass[conference]{IEEEtran}
\IEEEoverridecommandlockouts
\usepackage{cite}
\usepackage{amsmath,amssymb,amsfonts}
\usepackage{algorithm}
\usepackage{algorithmic}
\usepackage{graphicx}
\usepackage{textcomp}
\usepackage{xcolor}
\usepackage{arydshln}
\usepackage{hyperref}

\usepackage{color}
\usepackage{graphicx}
\usepackage{amsmath}
\usepackage{amsthm}
\usepackage{amsmath}
\usepackage{amssymb}
\usepackage{multirow}
\usepackage{booktabs}
\usepackage{subfigure}

\makeatletter
\newcommand{\ssymbol}[1]{$^{\@fnsymbol{#1}}$}
\makeatother

\def\BibTeX{{\rm B\kern-.05em{\sc i\kern-.025em b}\kern-.08em
    T\kern-.1667em\lower.7ex\hbox{E}\kern-.125emX}}
\begin{document}

\title{Improve Text Classification Accuracy with \\ Intent Information}

\author{
    Yifeng Xie
    \\
    \\
    School of Mathematics and Statistics, \\Guangdong University of Technology, China\\
    }

\maketitle

\begin{abstract}
Text classification, a core component of task-oriented dialogue systems,
attracts continuous research from both the research and industry community, and has resulted in tremendous progress.
However, existing method does not consider the use of label information, which may weaken the performance of text classification systems in some token-aware scenarios.
To address the problem, in this paper, 
we introduce the use of label information as label embedding for the task of text classification and achieve remarkable performance on benchmark dataset.
\end{abstract}

\begin{IEEEkeywords}
Text Classification, Label Information, Label Embedding, Pre-trained Language Model.
\end{IEEEkeywords}

\section{Introduction}
In recent years, goal-oriented dialogue systems have been widely applied in intelligent voice assistant, 
e.g., Apple Siri, Amazon Alexa, where intent classification technology plays a crucial part.
Given input utterance in natural language, the intent classification module aims to detect the user's intent~\cite{huang2019lattice,huang2020contextualSLU,zhou2020pin,huang2021sentiment}.
Previous works have been proposed for better understanding the semantic of the utterance~\cite{huang2022slt,huang2020FLSLU,chen2022han,chen2022bilinear}.
A simple example of intent classification is shown in Figure~\ref{tab:example}.

Despite the impressive results, most of the existing methods focus only on semantic understanding of utterances and are directly trained on the input utterances only.
Because the design and training process of models are noise-agnostic, it is difﬁcult for the model to adapt the knowledge learned from fixed and limited training set to the unknown open-domain user inputs directly, where the two datasets may be completely different in terms of data distributions. It means that a well-trained model on the ideal dataset may have poor performance on the open-domain user inputs, which indicates the current models suffer from poor robustness.
In short, we argue that for the intent classification task, studies used to model on a given dataset barely on the sentences are not friendly to the model robustness. Intuitively, learning from label information at the same time will improve the performance and robustness of the model, compared to the sentence-only training. Thus, it is reasonable to combine label information and text information for modeling in text classification task.

Conventional goal-oriented dialogue system mainly train intent classification model with cross entropy by treating the label as on-hot embedding.
However, in real application scenarios, the input sentence may contain many errors, like insertion error, deleting error, subtitution error.
In this manner, the overall sentence representation is hard to be understood and easy to lead to wrong prediction.

In addition, existing text classification approaches only consider the utterances in the coarse granularity level, which may less the possibility of the model to explore relationship between token-level information and label information.
For example, in Figure 1, if there is a token ``When" in the input sentence, then the ``NUM" intent is more likely to be recognized as high correlation with it, while the ``LOC" intent does not.

%%%%%%%%%%%%%%%%%%%%%%%%%%%%%%%%%%%%%%%%%%%%%%%%%%%%%%%%%%%%%%%%%%%
\begin{table}
\caption{Text classification example on TREC6 dataset.}
\begin{center}
\begin{tabular}{|l|c|c|}
\hline 
\textbf{Intent} & \textbf{Text} \\ \hline
NUM & When was the first liver transplant \\ \hline
LOC & What is the largest city in the world \\ \hline
ENTY & What do you call a newborn kangaroo \\ \hline
DESC & What does cc in engines mean \\ \hline
\end{tabular}
\end{center}
\label{tab:example}
\end{table}
%%%%%%%%%%%%%%%%%%%%%%%%%%%%%%%%%%%%%%%%%%%%%%%%%%%%%%%%%%%%%%%%%%%

To solve the above issues, in this paper, we propose to joint model label information and text information for the text classification task, specially, we consider the goal-oriented dialogue system scene that the label is an intent.

\section{Related Work}
\subsection{Text Classification}
Text classification is a common task in natural language processing (NLP), where the goal is to assign a label or category to a given piece of text. It is often used in applications such as sentiment analysis, where the goal is to predict whether a piece of text is positive or negative, or topic classification, where the goal is to identify the topic of a text. In related works, text classification has been extensively studied and many different approaches have been proposed. Some common approaches include using bag-of-words models, which represent the text as a collection of individual words, and using word embedding models, which capture the meaning and context of words in a continuous numerical space. More recently, transformer models like BERT have been shown to achieve remarkable performance on text classification tasks. Overall, text classification is an important and active area of research in NLP, with many different approaches and techniques being developed and studied.

\begin{figure*}[htbp]
\centerline{\includegraphics[width=0.30\linewidth]{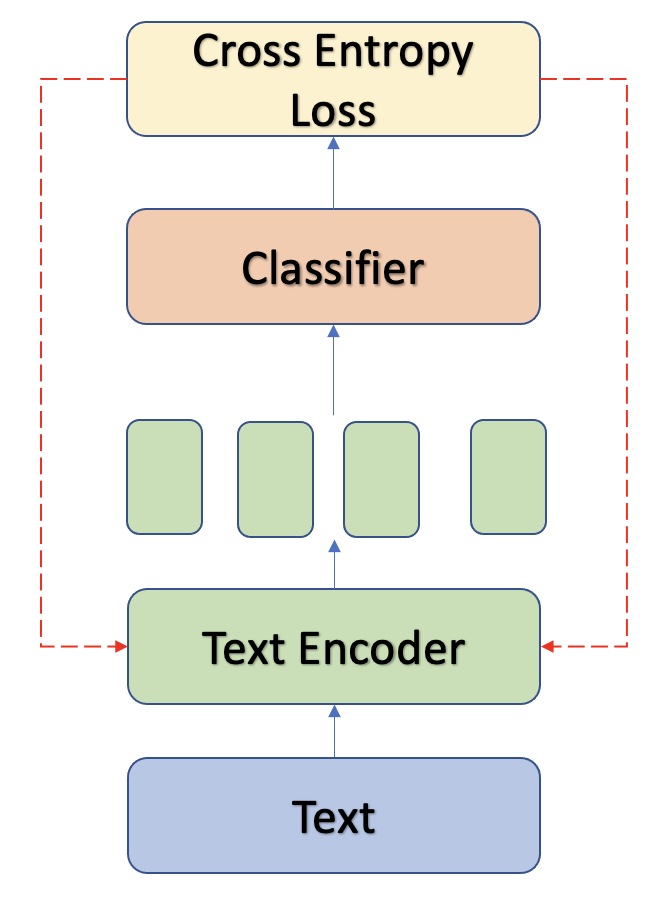}}
\caption{Conventional PLM-based text classification framework.}
\label{fig:orig}
\end{figure*}

\subsection{Pre-trained Language Model for Text Classification}
Due to the powerful text representation provided by the BERT-like Pre-trained Language Model (PLM)~\cite{Devlin2019BERT,liu2019roberta,hou2020dynabert,huang2021ghostbert}, previous works adopted BERT-like PLM to the text classification task, achieving remarkable success~\cite{wang2018labelEmbedding,Chen2019jointbert}.
BERT stands for "Bidirectional Encoder Representations from Transformers." BERT is a type of transformer model, a neural network architecture that uses self-attention mechanisms to process input data. BERT is trained to predict missing words in a sentence, which allows it to understand the context of words in a sentence and use that understanding to perform various NLP tasks such as sentiment analysis and question answering.

\subsection{Label Embedding}
\cite{wang2018labelEmbedding} proposed to joint train the label embedding and text embedding for the text classification task. However, their work does not consider the problem of noise text, as well as explore the ability of powerful PLM. In contrast, they inject the label information to the text encoding process by carefully designing the merging module.
Recent works~\cite{Chen2022DualCL} explored the effect of label information
by simultaneously learning the features of input samples and the parameters of classifiers in the same space,
and showed that label information is good for the intent classification task.
%However, their method introduce more data and cause long time training.

%Contrastive learning has achieved remarkable success in representation learning via self-supervision in unsupervised settings. However, effectively adapting contrastive learning to supervised learning tasks remains as a challenge in practice.

\section{Method}
\begin{figure*}[htbp]
\centerline{\includegraphics[width=0.50\linewidth]{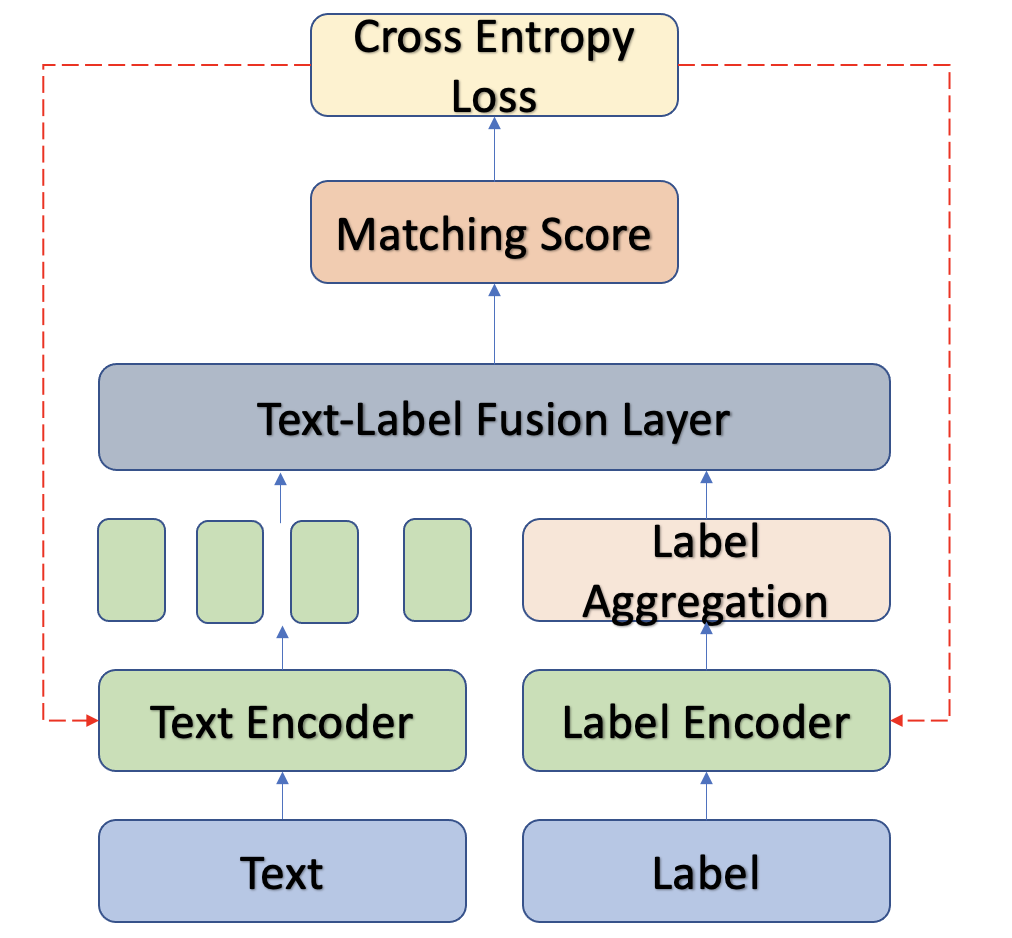}}
\caption{The general architecture of the proposed joint label method. 
Given text T and label L, the model first encode each of them and then fuse the two representation and treat the 
text classification problem as a matching problem.}
\label{fig:framework}
\end{figure*}

In this section, we will describe the proposed framework in detail.
Figure~\ref{fig:framework} shows the overall architecture of the model.

\subsection{Label-Sentence Co-Attention Model}
\subsubsection{Text Encoder}
Our sentence encoders are based on the pre-trained language model, RoBERTa.
RoBERTa~\cite{liu2019roberta} stands for Robustly Optimized BERT Pre-training Approach.
Give text input with $N$ tokens, RoBERTa then encode the text and output $N * d$ text representation, where $d$ is the embedding dimension.

\subsubsection{Label Encoder}
The label encoder aims to encode the label information after it is converted to label embedding, which is a
learnable embedding that takes a label in text format as input to RoBERTa.
Note that we share the parameters of text encoder and the label encoder.

%\subsubsection{Fusion Module}
\subsubsection{Text-Label Fusion Layer}
There are many ways to fuse two embeddings in NLP. One common method is to concatenate the two vectors, creating a single vector that contains both the text and label information. This vector can then be fed into a neural network for further processing and analysis. Another method is to use a weighted average of the two vectors, where the weights reflect the relative importance of the text and label information. This allows the model to give more or less emphasis to each type of information depending on the task at hand.
%There are many other possible ways to fuse text and label embedding, and the best approach will depend on the specific details of the NLP task and the data being used.
Inspired by CLIP~\cite{Radford2021clip}, we fuse the text information and label information by employing dot-product on the output representation of text encoder and label information.\nocite{huang2021audioMRC,liu2019align}

\subsection{Classifier}
BERT preprocessed inputs are fed into the BERT classifier, which processes the data and produces an output prediction.
In the case of a text classification task, the output prediction would be the predicted label or category for the input text.

\subsection{Training Objective}
In this work, cross entropy loss is used to optimize the probability from the predicted label to the ground truth label.
It can be defined as $ loss(y_{true}, y_{pred}) = - \frac{1}{n} \sum_{i=1}^{n} \left( y_{true} \log(y_{pred}) + (1 - y_{true}) \log(1 - y_{pred}) \right) $.

\section{Experiment}

\subsection{Datasets}
% done.
To evaluate the efﬁciency of our proposed method, we conduct experiments on benchmark dataset TREC and ATIS.
The detail of the datasets can be seen in Table~\ref{tab:dataset}.
Datasets used in our paper follows the same format and partition as in~\cite{Chang2022spokencse}.
\nocite{Bastianelli2020slurp,Prashanth2019atis2}
Intent detection accuracy is treated as evaluation metric in the experiments.

\begin{table}[ht]
\caption{Dataset statistics on TREC6 and ATIS.}
\centering
\resizebox{0.9\linewidth}{!}{
\begin{tabular}{lcccc}
\toprule
Dataset & \#Class & Avg. Length &  Train & Test \\ \midrule
TREC &  6   &  8.89 & 5,452 & 500  \\ 
ATIS &  22   &  11.14 & 4,978 & 893  \\ 
\bottomrule
\end{tabular}}
\label{tab:dataset}
\end{table}

\paragraph{TREC6}
The TREC6~\cite{hovy2001trec1,li2002trec2} dataset is dataset for question classification.
TREC6 consists of 6 Intents \textit{ABBR, ENTY, DESC, HUM, LOC, NUM}.
It has 5,452 training examples and 500 test examples.

\paragraph{ATIS}
The Airline Travel Information System (ATIS) dataset is a benchmark dataset commonly used to evaluate the performance of different models on a NLU task.
The ATIS dataset contains a large collection of sentences and questions related to flight reservations, along with the correct intent for each sentence.
For example, a sentence in the ATIS dataset might be "I would like to book a flight from New York to Los Angeles" and the corresponding label might be "book flight."

\subsection{Experimental Settings}
We use RoBERTa-base~\cite{liu2019roberta} as the text encoder for all experiments.
We follow~\cite{Chang2022spokencse} for the data split and data preprocessing.
To give the PLM a better initialization, we take the pre-trained RoBERTa checkpoint from~\cite{Chang2022spokencse} as our text and label encoder checkpoint for all experiments.
The training batch size is selected from [32, 64]. For each experimental setting, we train the model with 10 epoch.

\subsection{Results}
\begin{table}[htbp]
	\caption{Performance on the benchmark datasets under different pre-trained language models.}
	\begin{center}
	\resizebox{0.88\linewidth}{!}{
	\begin{tabular}{l  c c c c c c}
		\toprule
		   & Label Embeddings & Fusion Methods  & TREC6 & ATIS  \\ [0.5ex] 
		\midrule
		 & No  & No & 84.5 &  94.3  \\
	      & Yes  & Add & 84.7 & 94.6 \\
	      & Yes  & Dot Product & \textbf{85.3} & \textbf{95.1} \\
    \bottomrule
	\end{tabular}
	}
	\end{center}
	\label{tab:Main}
\end{table}

Table~\ref{tab:Main} shows the main results of the experiments.
We can see that after adding label information, the fusion model~\ref{fig:framework} perform better than the model without explicit label information~\ref{fig:orig}.
The results indicate that the model that incorporates label information (the fusion model) performs better than the model that does not (the original model). This suggests that incorporating label information can be beneficial for text classification tasks, as it can help the model to better understand the context and meaning of the text and make more accurate predictions. These results are consistent with many previous studies in natural language processing, which have shown that incorporating label information can improve the performance of text classification models.

\section{Conclusion}
%As can be seen from main results, our method is more accurate and general.
The label information introduced by this method enables the model to perform better text classification.
This is a common approach in natural language processing (NLP) and can be very effective at improving the accuracy of text classification models.
By incorporating label information, the model is able to better understand the context and meaning of the text, which can help it to make more accurate predictions. Additionally, by explicitly modeling the interaction between the label and the text, the model can learn more about the token-level information, such as the individual words and their relationships to one another, which can further improve its performance on the text classification task. Overall, incorporating label information into a text classification model can be a powerful technique for improving its performance and enhancing its ability to understand natural language
%At the same time, this method can explicitly increase the cognitive ability of the text classification model for token-level information by introducing the interaction between label and text.

\bibliographystyle{plain}
%\bibliography{references}
\bibliography{acl2021}

\end{document}